\documentclass{article}
\usepackage{spconf,epsfig}
\usepackage{amsmath,amsfonts,amsthm}
\usepackage{algorithm}
\usepackage{algorithmic}
\usepackage{amssymb}
\usepackage{graphicx}
\usepackage{subfigure}
\usepackage{booktabs}
\usepackage{enumitem}
\usepackage{marvosym}
\def\etal{\emph{et al.}}
\let\OLDthebibliography\thebibliography
\renewcommand\thebibliography[1]{
  \OLDthebibliography{#1}
  \setlength{\parskip}{0pt}
  \setlength{\itemsep}{0pt plus 0.3ex}
}

\begin{document}\sloppy


\title{AugRmixAT: A Data Processing and Training Method for Improving Multiple Robustness and Generalization Performance}
%
\name{Xiaoliang Liu$^{\ast}$, Furao Shen $^{\ast}$\textsuperscript{\Letter} , Jian Zhao$^{\dagger}$, Changhai Nie$^{\ast}$}
\address{$^{\ast}$$^{\dagger}$National Key Laboratory for Novel Software Technology, Nanjing University;\\ $^{\ast}$Department of Computer Science and Technology, Nanjing University;\\ $^{\dagger}$ School of Electronic Science and Engineering, Nanjing University.\thanks{\textsuperscript{\Letter}Furao Shen is the corresponding author. Email: frshen@nju.edu.cn. This work was supported by the State Grid Corporation of China and its project number is 520950200009.}}
\maketitle

\begin{abstract}
Deep neural networks are powerful, but they also have shortcomings such as their sensitivity to adversarial examples, noise, blur, occlusion, etc. Moreover, ensuring the reliability and robustness of deep neural network models is crucial for their application in safety-critical areas. Much previous work has been proposed to improve specific robustness. However, we find that the specific robustness is often improved at the sacrifice of the additional robustness or generalization ability of the neural network model. In particular, adversarial training methods significantly hurt the generalization performance on unperturbed data when improving adversarial robustness. In this paper, we propose a new data processing and training method, called AugRmixAT, which can simultaneously improve the generalization ability and multiple robustness of neural network models. Finally, we validate the effectiveness of AugRmixAT on the CIFAR-10/100 and Tiny-ImageNet datasets. The experiments demonstrate that AugRmixAT can improve the model's generalization performance while enhancing the white-box robustness, black-box robustness, common corruption robustness, and partial occlusion robustness.
\end{abstract}

\section{Introduction}
\label{sec:intro}
Deep neural networks have achieved remarkable success in a variety of fields and have been widely used in areas where reliability and security are critical, such as medical image processing~\cite{shen2017deep}, autonomous driving~\cite{grigorescu2020survey}, and face recognition~\cite{sharif2017adversarial}. Unfortunately, recent studies~\cite{fgsm,madry2018towards} have shown that artificially adding an imperceptible adversarial perturbation to the input image can significantly reduce the recognition ability of the neural network or guide the neural network to identify it as the characteristic wrong target. In addition to artificially designed adversarial examples, there are various common corruptions~\cite{hendrycks2019benchmarking} and occlusions~\cite{devries2017improved} in real-world environments that also affect the robustness and reliability of neural networks. 

Many methods~\cite{fgsm,madry2018towards,dabouei2020exploiting,zhang2019theoretically,lamb2019interpolated} have recently been proposed for defending against adversarial attacks. Among such defense methods, adversarial training has proven to be one of the most promising methods~\cite{madry2018towards,zhang2019theoretically,lamb2019interpolated}. However, adversarial training also has a huge drawback in that it drastically reduces the generalization ability of neural networks on the original data~\cite{madry2018towards,zhang2019theoretically}. Furthermore, we find that adversarial training is similarly detrimental to occlusion robustness. Previous studies~\cite{schmidt2018adversarially,engstrom2019exploring} have also shown that improving one specific robustness is not necessarily beneficial or even harmful to another specific robustness. In our experiments, we can also discover that CutMix~\cite{cutmix} can effectively enhance the occlusion robustness but is detrimental to noise robustness and adversarial robustness. However, for security-sensitive applications in practice, we cannot consider only a single specific robustness, but multiple aspects of robustness and generalization performance of neural network models. 

To address the above issues, we propose AugRmixAT, a new data processing and training method that can simultaneously improve the multiple robustness and generalization performance of neural network models. AugRmixAT utilizes traditional data augmentation, mixed data augmentation~\cite{mixup,cutmix,harris2020fmix,qin2020resizemix} between different samples, and data augmentation with added adversarial perturbation to process and generate multiple different sets of augmented data. To ensure the generalization performance on clean data (standard test data), AugRmixAT uses both soft cross-entropy and Jensen-Shannon divergence~\cite{endres2003new} consistent loss to train multiple sets of augmented data in a surrogate manner. Finally, we experimented on CIFAR-10/100 and Tiny-ImageNet~\cite{chrabaszcz2017downsampled} and showed that AugRmixAT can simultaneously improve white-box robustness, black-box robustness, 19 common corruption robustness on CIFAR-10/100, 15 common corruption robustness on Tiny-ImageNet, partial occlusion robustness, and generalization performance on clean data. 

\begin{figure*}[t]
	\centering
	\includegraphics[width=0.7\linewidth]{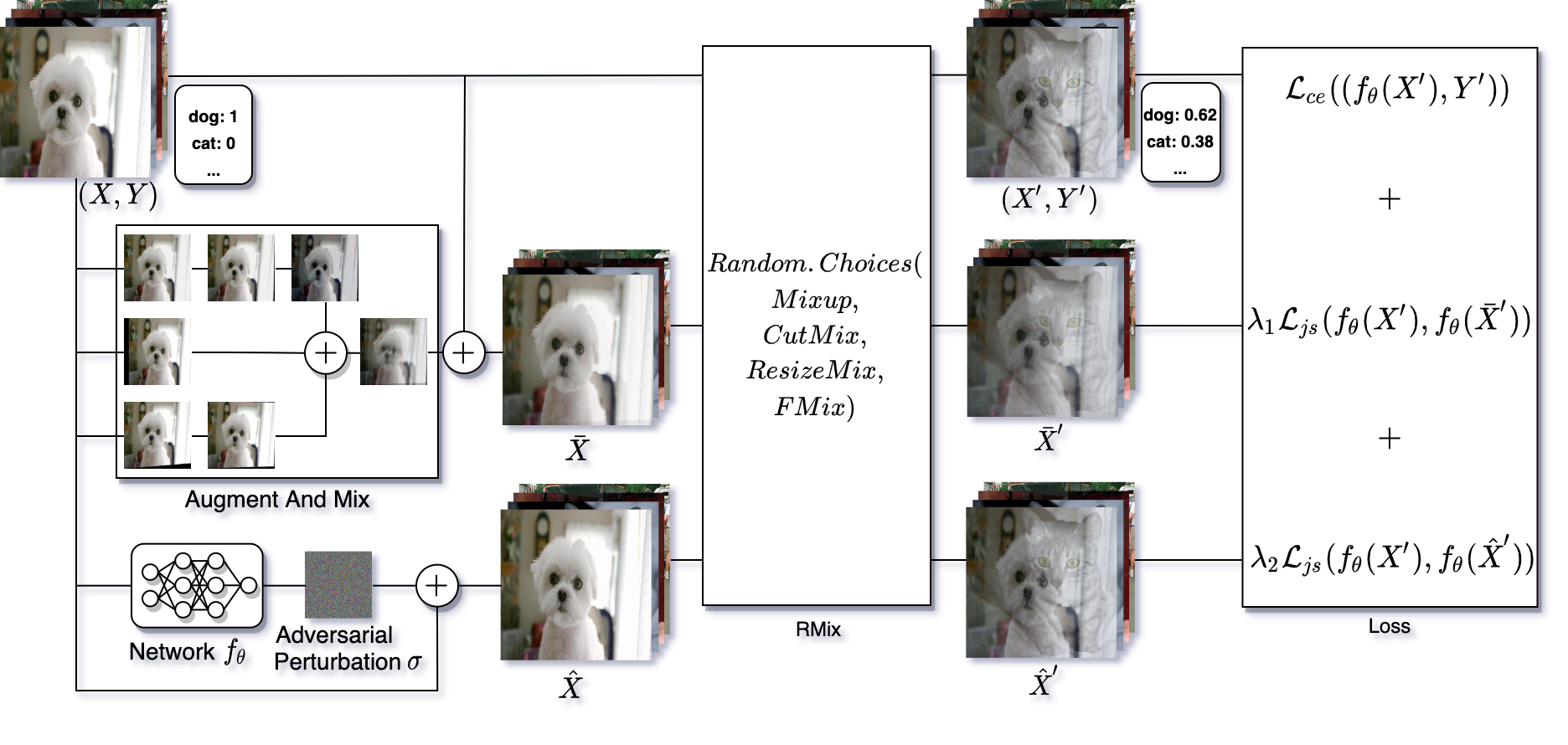}
	\caption{An example of AugRmixAT.}
	\label{fig:augrmixat}
\end{figure*}

\section{Related Work}
\label{sec:related_work}

\noindent\textbf{Data Augmentation.}  
Data augmentation is a very practical and powerful technique to increase the diversity of training datasets, enhance the generalization ability of neural networks and prevent overfitting~\cite{bishop2006pattern}. For instance, some of the most commonly used data augmentations in computer vision are geometric transformations, flipping, color modification, cropping, rotation, translation, noise injection and random erasing~\cite{shorten2019survey}. Recently, mixed sample data augmentation methods have gained tremendous attention and a series of mixed sample data augmentation methods~\cite{mixup,cutmix,harris2020fmix,qin2020resizemix} have been proposed. Mixup~\cite{mixup} is the first proposed mixed sample data augmentation that mixes two different samples in a convex combination to generate a new training sample and corresponding label. Combining the ideas of Mixup and Cutout~\cite{devries2017improved}, CutMix~\cite{cutmix} uses cutting and pasting patches between training images for mixing, and ground truth labels are also proportionally mixed with patch regions. Fmix~\cite{harris2020fmix} uses a random binary mask obtained by applying a threshold to low-frequency images sampled from Fourier space to further improve the shape of CutMix mixed patches. To solve the problem of label misallocation and object information missing in CutMix, ResizeMix~\cite{qin2020resizemix} mixes training data by directly resizing the source image to a small patch and then pasting it on another image. AugMix~\cite{hendrycks2020augmix} is proposed to improve both the generalization performance and the corruption robustness by mixing common data augmentation.  

\noindent\textbf{Adversarial Training (AT).} 
Adversarial training, which augments training dataset with adversarial examples, is one of the most effective methods of defending against adversarial attacks~\cite{fgsm,madry2018towards,zhang2019theoretically,lamb2019interpolated}. Therefore, we can also consider adversarial training as a data augmentation technique. Goodfellow~\etal~\cite{fgsm} proposed the Fast Gradient Sign Method (FGSM), which is a simple and fast method to generate adversarial examples for adversarial training. Projected Gradient Descent (PGD)~\cite{madry2018towards} adversarial training leverages the PGD attack to generate adversarial examples and trains only with the adversarial examples. Zhang~\etal~\cite{zhang2019theoretically} proposed TRADES to specifically maximize the trade-off of adversarial training between adversarial robustness and standard accuracy. Lamb~\etal~\cite{lamb2019interpolated} proposed Interpolated Adversarial Training(IAT), which trains on interpolations of adversarial examples along with interpolations of unperturbed examples and improves adversarial robustness without sacrificing too much standard accuracy.

\section{AugRmixAT}
\label{sec:method}
Previous data augmentation~\cite{mixup,cutmix,hendrycks2020augmix} and adversarial training~\cite{madry2018towards,zhang2019theoretically,lamb2019interpolated} methods can effectively improve specific robustness or generalization performance, but they are difficult to improve multiple robustness and generalization abilities of deep neural network models simultaneously. In particular, most adversarial training~\cite{madry2018towards,zhang2019theoretically,lamb2019interpolated} tends to sacrifice standard accuracy when enhancing adversarial robustness. AugRmixAT is an image data processing and training method that can simultaneously improve multiple robustness and generalization performance of models and is easy to slot into existing training pipelines. Figure~\ref{fig:augrmixat} shows an example of AugRmixAT. First, a batch of input images $X$ is used to generate data $\bar{X}$ by ``Augment And Mix'' data augmentation and to generate adversarial samples $\hat{X}$ by adding adversarial perturbations, respectively. Next, $X$, $\bar{X}$, and $\hat{X}$ are processed with mixed sample data augmentation to generate $X'$, $\bar{X}'$, and $\hat{X}'$. The corresponding mixed labels $Y'$ are also generated using the labels $Y$. Finally, we use a soft cross-entropy loss $\mathcal{L}_{ce}$ and Jensen-Shannon divergence consistent loss $\mathcal{L}_{js}$ to train $X'$, $\bar{X}'$, and $\hat{X}'$.


\noindent\textbf{Augment And Mix.} We use the same ``Augment And Mix'' operation as AugMix~\cite{hendrycks2020augmix}. The ``Augment And Mix'' operation starts by randomly selecting multiple augmentations from the base augmentation set to form multiple augmentation chains and producing multiple augmentation samples through the augmentation chains. Then, multiple augmentation samples are mixed through a random convex combination sampled from a Dirichlet($\alpha,...,\alpha$) distribution. Finally, we combine this mixed sample with the original sample through a second random convex combination sampled from a Bata($\alpha,\alpha$) distribution. In the experiment, we put $\alpha$ to 1 and the number of augmentation chains to 3. Each augmentation chain consists of 1 to 3 random base augmentation operations. Our base data augmentation set contains $autocontrast$, $equalize$, $rotate$, $solarize$, $shear$, $translate$. 

\noindent\textbf{Adversarial examples.} We apply PGD~\cite{madry2018towards} adversarial attacks to generate adversarial examples, which can be expressed as 
\begin{equation}
\begin{aligned}
	 &\hat{x}^{0}= x + 0.001 \cdot \mathcal{N}(\mathbf{0},\mathbf{I}), \\	
	&\hat{x}^{t+1} = \prod_{\mathcal{B}_{\infty}(x,\epsilon)}(\eta \mathbf{sign}(\nabla_{\hat{x}^{t}} \mathcal{L}_{kl} (f_\theta(x), f_\theta(\hat{x}^{t})))+\hat{x}^{t}),
\end{aligned}
\end{equation}
where $\mathcal{N}(\mathbf{0},\mathbf{I})$ is the Gaussian distribution function with zero mean and identity variance, $\epsilon$ is the adversarial perturbation budget, $\eta$ is the perturbation step size, $\mathcal{B}_{\infty}(x,\epsilon)$ represents a neighborhood of $x:\{x^{adv}:||x-x^{adv}||_{\infty}\leq \epsilon\}$, $\mathbf{sign}(\cdot)$ is the sign function, $\mathcal{L}_{kl}(\cdot)$ is the KL divergence loss function, $f_\theta$ denotes the neural network with parameters $\theta$.

\noindent\textbf{Mixed Sample Data Augmentation.} 
To further obtain more diverse training data, we simultaneously perform the same mixed sample data augmentation operation on the original images $X$, the ``Augment And Mix'' enhanced images $\bar{X}$, and the adversarial examples $\hat{X}$. In our work, we integrate multiple mixed sample data augmentation~\cite{madry2018towards,zhang2019theoretically,lamb2019interpolated} in a randomly chosen manner. Our mixed sample data augmentation operates as follows,
\begin{equation}
\begin{aligned}
	& rand\_index = Randperm(batch\_size),\\
	& \begin{aligned}Mix = Random.Choices(Mixup,CutMix,\\
	 ResizeMix,FMix),\end{aligned} \\
	& X' = Mix(X,X[rand\_index]),\\
	& \bar{X}' = Mix(\bar{X},\bar{X}[rand\_index]),\\
	& \hat{X}' = Mix(\hat{X},\hat{X}[rand\_index]),\\
	& Y' = \gamma Y + (1-\gamma) Y[rand\_index],
\end{aligned}
\end{equation}
where $Randperm(\cdot)$ is the random permutation function, $Random.Choices(\cdot)$ is the random choice function, $\gamma$ is the corresponding mixing ratio. 

\noindent\textbf{Loss Function.} To ensure the generalization ability of the model and to improve the robustness of the model, we use the Jensen-Shannon divergence consistent loss function to train the mixed ``Augment And Mix'' enhanced data $\bar{X}'$ and mixed adversarial examples $\hat{X}'$ in a surrogate manner. Our loss function is defined as
\begin{equation}
\begin{aligned}
	\mathcal{L} = \mathcal{L}_{ce}(f_\theta(X'),Y') + \lambda_1 \mathcal{L}_{js}(f_\theta(X'),f_\theta(\bar{X}'))+&\\ \lambda_2\mathcal{L}_{js}(f_\theta(X'),f_\theta(\hat{X}')),&
\end{aligned}
\end{equation}
where $\lambda_1$ and $\lambda_2$ are two regularization hyperparameters. The detailed algorithm is described in Algorithm Block~\ref{alg:aug_rmix_at}.
\begin{algorithm}[tbh]
\normalsize
	\caption{AugRmixAT algorithm}	
	\label{alg:aug_rmix_at}
	\begin{algorithmic}[1]
		\REQUIRE{Training dataset $D$, Perturbation $\epsilon$, Perturbation step size $\eta$, Number of iterations $T$, Neural network parameters $\theta$, Neural network $f_\theta$}
		\REQUIRE{$\boldsymbol{AugmentAndMix}$ is the data augmentation and mixing function for the samples themselves, $\boldsymbol{RMix}$ is the mixed data augmentation function between different samples}, $\mathcal{N}(\mathbf{0},\mathbf{I})$ is the Gaussian distribution function with zero mean and identity variance
		\REQUIRE{KL divergence loss function $\mathcal{L}_{kl}$, Jensen-Shannon divergence loss function $\mathcal{L}_{js}$, Soft cross-entropy loss function $\mathcal{L}_{ce}$}
		
		\REPEAT
			\STATE{Read mini-batch $ (X,Y)=\{(x_i,y_i),...,(x_m,y_m)\}$ from training dataset $D$}
			\FOR{$i=1,...,m$ (in parallel)}
			\STATE $\bar{x}_i \leftarrow \boldsymbol{AugmentAndMix}(x_i)$
			\STATE $\hat{x}_i \leftarrow x_i + 0.001 \cdot \mathcal{N}(\mathbf{0},\mathbf{I})$ 
			\FOR{$t=1,...,T$}
			\STATE{\small $ 
			 \hat{x}_i\leftarrow\! \prod\limits_{\mathcal{B}_{\infty}(x_i,\epsilon)}(\eta \mathbf{sign}(\nabla_{\hat{x}_i} \mathcal{L}_{kl} (f_\theta(x_i),f_\theta(\hat{x}_i)))+\hat{x}_i)$}
			\ENDFOR 
			\ENDFOR
			\STATE $X',\bar{X}',\hat{X}',Y' \leftarrow \boldsymbol{RMix}(X,\bar{X},\hat{X},Y)$, where $\bar{X}=\{\bar{x}_i,...,\bar{x}_m\}$, $\hat{X}=\{\hat{x}_i,...,\hat{x}_m\}$
			\STATE $
			\mathcal{L} = \frac{1}{m} \sum\limits_{i=1}^m (\mathcal{L}_{ce}(f_\theta(x'_i),y'_i)+ \lambda_1 \mathcal{L}_{js}(f_\theta(x'_i),f_\theta(\bar{x}'_i))+ \lambda_2\mathcal{L}_{js}(f_\theta(x'_i),f_\theta(\hat{x}'_i)))$
			\STATE $\theta \leftarrow \theta-\beta \nabla_\theta\mathcal{L}$
		\UNTIL{training converged}
	\end{algorithmic}
\end{algorithm}

\begin{table*}[th]
\centering
\caption{Corruption Error (CE,\%), and mCE (\%) values on CIFAR-10-C. The mCE value is calculated by averaging all 19 CE values.}
\label{tab:cifar10_c}
\scalebox{0.5}{

\begin{tabular}{@{}lcccc|ccccc|ccccc|ccccc|c@{}}
\toprule
   & \multicolumn{4}{c|}{Noise} & \multicolumn{5}{c|}{Blur} & \multicolumn{5}{c|}{Weather} & \multicolumn{5}{c|}{Digital} &  \\
Model  & Gauss. & Shot & Impulse & Speckle & Defocus & Glass & Motion & Zoom & Gauss. & Snow & Frost & Fog & Spatter & Bright & Contrast & Elastic & Pixel & Saturate & JPEG  & mCE \\ \midrule
Standard   & 52.64 & 39.49 & 48.86 & 35.44  & 14.13 & 43.36 & 18.18 & 16.10 & 20.71 & 14.29 & 16.96 & 9.13 & 14.24 & 5.11  & 18.80  & 14.00 & 22.41 & 6.73 & 20.83 & 22.71  \\
Mixup   & 32.40  &29.68 &33.39 & 27.11 &11.03  &32.4  & 13.61 & 14.13 & 22.27 &8.16  & 7.52 & 7.83& 6.53 &4.06  &  21.97& 11.38  & 22.01 & 5.56 & 17.23 &17.65  \\
CutMix  & 79.40 &  68.68 & 61.24 & 66.91 &15.87  &44.70  & 19.06 &20.45 & 28.56  &12.25  & 18.48 & 8.16& 7.85 & 4.84 &13.36  & 14.74 &30.25 & 7.04  & 29.93 &29.04  \\
AugMix   & 16.38  & {11.83} & 9.83 & \bf{10.41}    &4.17     &19.17  & 5.46 & 5.22 & 4.63  &7.67  &7.66   &5.65 & 5.18&3.85    & \bf{5.81}      &8.08     &\bf{10.29} & 5.54  &11.50  &8.33  \\ \midrule
PGDAT   &18.29  & 16.98 &30.43 & 17.06  &  18.49 & 20.33 & 22.84 &  19.59 & 21.06 &19.46  & 22.85 &  41.16& 17.97 & 16.50 & 55.51 & 19.44 & 15.34 & 16.81 & 15.41 &22.40  \\
TRADES ($1/\lambda=1$)   & 17.75 &16.60  &  27.94 &16.73 &18.15  & 20.35 & 22.38 & 19.40 & 20.59 & 19.44 & 23.75 & 40.06 &  17.53 & 16.57 & 54.54 & 19.35 &15.10 & 16.63  &15.04  & 22.00 \\
TRADES ($1/\lambda=6$)   &19.56  &18.62  & 29.35 &19.04 & 19.82  & 21.51  &24.30  &20.81 & 22.00  & 21.08 & 25.46 & 40.96& 19.44 & 18.41 & 56.47 &20.91  & 16.84 & 18.41 & 16.85 &23.68  \\
IAT  & 15.48 & 13.78 &27.68 & 14.08  &8.73  & \bf{15.18} &10.90  & 10.33 & 13.61 &10.05  &9.23  &10.29& 7.59  & 6.63 & 19.30 & 9.67 & 11.73 & 7.59 & {11.32} &12.27  \\ \midrule
AugRmixAT-1-1 & \bf{13.71} & \bf{11.19 } & \bf{7.33} & 10.54  & \bf{3.10}  & 18.56 & \bf{5.45} & \bf{4.11} & \bf{3.82} & \bf{5.45} & \bf{6.18}  & \bf{4.26}& \bf{3.03} & \bf{2.97} &  8.05 & \bf{5.98} & 15.26 & \bf{3.93} & \bf{10.73} & \bf{7.56} \\
AugRmixAT-1-32 & 14.69 & 13.67 &17.46 & 13.76  &15.09  &17.87  &18.56  & 16.11 & 16.91 &16.70  &  19.46&32.36& 14.39  & 14.07 & 49.29 &16.46  & 12.97 & 14.39 & 13.21 &18.25  \\ \bottomrule
\end{tabular}
}
\end{table*}
\begin{table}[th]
\centering
\caption{White-box Top1 robust accuracy (\%) and Clean Top1 accuracy (\%) on CIFAR-10.}
\label{tab:cifar10_w}
\scalebox{0.6}{
\begin{tabular}{@{}lc|cccc@{}}
\toprule
 &  & \multicolumn{4}{c}{White-box attacks}  \\
Model & Clean & FGSM & PGD10 & PGD20 & CW20  \\ \midrule
Standard & 96.06 & 56.30 & 5.10 & 0.76 & 0.22   \\
Mixup & 97.12 & 65.71 & 14.21 & 2.99 & 0.99   \\
CutMix & 96.90 & 41.40 & 4.37 & 1.63 & 0.55   \\
AugMix & 96.49 & 53.38 & 4.01 & 0.23 & 0.09   \\ \midrule
PGDAT & 87.06 & 59.04 & 62.13 & 51.11 & 50.81   \\
TRADES ($1/\lambda=1$) & 87.37  & 59.21 & 61.46 & 50.50 & 50.27  \\
TRADES ($1/\lambda=6$) & 85.35 & 59.37 & 61.84 & 51.79 & 52.70   \\
IAT & 95.27 & 86.59 & 63.39 & 53.49 & 54.15  \\ \midrule
AugRmixAT-1-1 & \bf{98.17} & \bf{88.01} & {66.22} & 53.21 & 54.33   \\
AugRmixAT-1-32 & 89.14 & 65.88 & \bf{67.81} & \bf{57.50} & \bf{56.47} \\ \bottomrule
\end{tabular}
}
\end{table}

\begin{table}[th]
\centering
\caption{Black-box Top1 robust accuracy (\%) on CIFAR-10.}
\label{tab:cifar10_black}
\scalebox{0.7}{
\begin{tabular}{@{}lcccc@{}}
\toprule
Defense & \multicolumn{4}{c}{Attack model} \\ \cmidrule(l){2-5} 
model & Standard & PGDAT & TRADES & IAT \\ \midrule
PGDAT 			& 86.40 & - & 69.38 & 77.44 \\
TRADES ($1/\lambda=1$)& 86.60 & 68.69 & - & 76.87 \\
IAT	& 89.81 & 77.01 & 76.70 & - \\ \midrule
AugRmixAT-1-1   & \bf{91.96} & \bf{87.37} & \bf{87.53} & \bf{83.57}\\
AugRmixAT-1-32  & 88.49 & 75.20 & 74.69 & 79.60 \\
 \bottomrule
\end{tabular}
}
\end{table}

\begin{table}[th]
\centering
\caption{Top1 robust accuracy (\%) of untargeted partial occlusion and Top2 robust accuracy (\%) of targeted partial occlusion on CIFAR-10.}
\label{tab:cifar10_occ}
\scalebox{0.7}{
\begin{tabular}{@{}lccc@{}}
\toprule
Model  & Untargeted & Targeted & Mean   \\ \midrule
Standard   & 77.58 & 76.41 &  77.00   \\
Mixup   & 81.24 & 82.69 & 81.97     \\
CutMix   & 90.95 & 92.59 &  91.77    \\
AugMix   & 78.47 & 77.60 & 78.03    \\ \midrule
PGDAT   & 56.49 & 66.22 & 61.36     \\
TRADES ($1/\lambda=1$)   & 57.77 & 63.14 & 58.95    \\
TRADES ($1/\lambda=6$)  & 54.77 & 63.14 & 58.95    \\
IAT   & 69.63 & 78.87 & 74.25   \\ \midrule
AugRmixAT-1-1   & \bf{91.04} & \bf{93.66} & \bf{92.35}     \\
AugRmixAT-1-32  & 75.44 & 80.79 & 78.12   \\ \bottomrule
\end{tabular}
}
\end{table}

\section{Experiments}
\label{sec:experiments}
\subsection{Implementation Details}
We use the same neural network architecture as in previous works~\cite{madry2018towards,zhang2019theoretically}, i.e., WideResNet-34-10~\cite{zagoruyko2016wide}, for experiments on CIFAR-10/100 and PreAct-ResNet18~\cite{he2016identity} for experiments on Tiny-ImageNet~\cite{chrabaszcz2017downsampled}. Except for the different neural network architecture, other settings and hyperparameters are the same for all datasets. We apply the momentum stochastic gradient descent optimizer on both CIFAR-10/100 and Tiny-ImageNet. The initial learning rate is set to 0.1 and decays with the cosine annealing schedule~\cite{2016SGDR}. We set the momentum as 0.9 and use the weight decay of $5 \times 10^{-4}$. The batch size for training is set to $ 128 \times 4 $ and the maximum number of epochs is set to 200. The following is the setting of our main comparison method in the experiment.
	 
	 \noindent\textbf{Standard}: The model trained on the original data does not use any data augmentation methods.
	
	 \noindent\textbf{Mixup, CutMix, AugMix}: The models trained using data augmentation methods Mixup~\cite{mixup}, CutMix~\cite{cutmix} and AugMix~\cite{hendrycks2020augmix} respectively.
	
	 \noindent\textbf{PGDAT, TRADES, IAT}: The models trained using PGD Adversarial Training (PGDAT)~\cite{madry2018towards}, TRADES~\cite{zhang2019theoretically}, and Interpolated Adversarial Training (IAT)~\cite{lamb2019interpolated} respectively, where the perturbation budget are set to 0.031, the perturbation step size are set to 0.007, and the number of iterations are set to 10. The way of combining examples in IAT is Mixup.
	
	\noindent\textbf{AugRmixAT-1-1, AugRmixAT-1-32}: The models trained using our proposed method, in which 
	the perturbation budget $\epsilon$, the perturbation step size $\eta$, and the number of iterations $T$ are set the same as in PGDAT, TRADES, and IAT. ``-1-1'' means $\lambda_1=1,\lambda_2=1$. ``-1-32'' means $\lambda_1=1,\lambda_2=32$.

Additionally, all experiments were implemented and evaluated on the PyTorch~\cite{paszke2017automatic} platform with four NVIDIA Tesla V100 GPUs.

\subsection{CIFAR-10}
\textbf{Evaluation on White-box Robustness.}  The results of the white-box robustness on CIFAR-10 are shown in Table~\ref{tab:cifar10_w}. We evaluate the robustness of all models against three types of white-box attacks for CIFAR-10, i.e., FGSM~\cite{fgsm}, PGD~\cite{madry2018towards}, and CW~\cite{carlini2017towards} (PGD with CW loss). For FGSM, we set the perturbation budget as 0.031. For PGD10, PGD20, and CW20, we set the perturbation budget to  0.031 and the perturbation step size to  0.003. PGD10 set the number of iterations as 10. PGD20 and CW20 set the number of iterations as 20.

We can see from Table~\ref{tab:cifar10_w} that all the compared adversarial training methods reduce the Clean accuracy, but the AugRmixAT-1-1 model trained by our method can improve the Clean accuracy. Moreover, it is 1.05\% higher than the Mixup. Under the FSGM attack, the AugRmixAT-1-1 model has the best robust accuracy. Under the attacks of PGD10, PGD20 and CW20 respectively, the AugRmixAT-1-32 model achieves the best robust accuracy. In particular, the robust accuracy rate on PGD20 of the AugRmixAT-1-32 model is 6.39\% higher than PGDAT, 5.71\% higher than TRADES ($1/\lambda=6$), and 4.01\% higher than IAT.

\textbf{Evaluation on Black-box Robustness.} We use transfer-based black-box attacks~\cite{papernot2017practical} to evaluate the black-box robustness of the model. We first use each trained model to construct adversarial examples by PGD and then apply these adversarial examples to other models and evaluate their performance. We set the perturbation budget as 0.031, the perturbation step size as 0.003, and the number of iterations as 10. The results of the black-box robustness on CIFAR-10 are reported in Table~\ref{tab:cifar10_black}. Again, the AugRmixAT-1-1 model trained by our method achieves higher robustness than the other models. 

\textbf{Evaluation on Common Corruptions Robustness.} 
We evaluate the robustness of various common corruptions on the CIFAR-10-C~\cite{hendrycks2019benchmarking}, which consists of 19 types of corruption. Moreover, each type of corruption has 5 levels of severity. 

Following prior works~\cite{hendrycks2019benchmarking,hendrycks2020augmix}, we adopt Corruption Error (CE)~\cite{hendrycks2019benchmarking} to measure the common corruption robustness and mCE denotes the mean Corruption Error of the 19 corruption. As shown in Table~\ref{tab:cifar10_c}, the AugRmixAT-1-1 model trained by our proposed method achieves the lowest CE on 15 of 19 common corruptions. Moreover, the mCE of AugRmixAT-1-1 is also the lowest, 0.77\% lower than Augmix, 15.15\% lower than Standard, 14.84\% lower than PGDAT, and 4.71\% lower than IAT. 

\textbf{Evaluation on Partial Occlusion Robustness.}
Compared to corruption and adversarial example attacks, partial occlusion should be more common. We use untargeted random partial occlusion and targeted random partial occlusion to evaluate the robustness of the model under partial occlusion attacks. Untargeted occlusion blocks are filled with 0 and targeted occlusion blocks are from other objects. For untargeted partial occlusion we used the Top1 robust accuracy metric and for targeted partial occlusion we used the Top2 robust accuracy. From Table~\ref{tab:cifar10_occ}, we can find that the previous adversarial training methods PGDAT, TRADES and IAT are difficult to defend against partial occlusion attacks and are even detrimental to the robustness of partial occlusion. In contrast, our method can not only effectively improve both targeted and untargeted occlusion robust accuracy, but also has a robust accuracy rate of 0.09\% higher than CutMix in the untargeted occlusion and 1.07\% higher than CutMix in the targeted occlusion. Furthermore, AugRmixAT-1-1 achieves the best performance under partial occlusion attacks, and far outperformed the models trained by other adversarial training methods.


\begin{table}[th]
\centering
\caption{Clean Top1 accuracy (\%) and robust accuracy (\%) under various attacks on CIFAR-100 and Tiny-ImageNet. }
\label{tab:cifar100}
\scalebox{0.54}{
\begin{tabular}{@{}lc|cccc|cc|cc@{}}
\toprule
\bf{CIFAR-100}
\\ \midrule
 &  & \multicolumn{4}{c|}{White-box attacks} & \multicolumn{2}{c|}{Black-box attacks} &  &  \\
Model & Clean & FGSM & PGD10 & PGD20 & CW20 & Standard & PGDAT & Corr & Occ \\ \midrule
Standard & 80.12 & 24.26  & 2.09 & 0.55 & 0.25 & - & 62.38 & 51.53 & 54.64 \\
Mixup & 82.53 & 40.19  & 0.52 & 0.05 & 0.00 & 56.44 & 68.03 & 58.49 & 60.73 \\
CutMix & 82.38 & 19.73 & 0.66 & 0.12 & 0.00 & 34.46 & 61.47 & 48.20 & 71.95 \\
AugMix & 80.21 & 19.54 &1.52 & 0.38 & 0.14  & 68.42 & 68.63 & 68.75 & 54.65 \\ \midrule
PGDAT & 61.16 & 30.76 & 33.95 & 26.48 & 25.43 & 60.79 & - & 48.72 & 35.12 \\
TRADES ($1/\lambda=1$) & 60.60 & 30.83 & 33.74 & 26.67 & 25.99 & 60.01 & 43.22 & 48.50 & 33.61 \\
TRADES ($1/\lambda=6$) & 56.30 & 29.63 & 32.64 & 25.78 & 25.53 & 55.75 & 41.83 & 45.63 & 30.79 \\
IAT & 75.79 & 58.07 & 33.79 & 24.95 & 20.27 & 68.94 & 53.55 & 61.49  & 50.15 \\ \midrule
AugRmixAT-1-1& \bf{84.83} & \bf{66.50} & {36.51} & 25.94 & 22.15 & \bf{76.89} & \bf{71.55} & \bf{71.31} & 7\bf{4.21}  \\
AugRmixAT-1-32 & 65.93  & 37.59 & \bf{38.49} & \bf{30.18} & \bf{28.50} & 65.05 & 49.58 & 55.20 & 56.38 \\ 
\toprule 
\bf{Tiny-ImageNet}
\\
\midrule
 &  & \multicolumn{4}{c|}{White-box attacks} & \multicolumn{2}{c|}{Black-box attacks} &  &  \\
Model & Clean & FGSM & PGD10 & PGD20 & CW20 & Standard & PGDAT & Corr & Occ \\ \midrule
Standard & 63.90 & 1.07 & 0.10 & 0.00 & 0.00 & - & 48.45& 24.31 & 49.34  \\
Mixup & 64.66 & 0.94 & 0.00 & 0.00 & 0.00 & 20.68 & 51.07 & 27.84 & 50.62 \\
CutMix & 67.78 & 2.75 & 0.02 & 0.00 & 0.00 & 13.80 & 53.44 & 24.63 & 62.26 \\
AugMix & 62.46 & 2.83 & 0.16 & 0.03 & 0.00 & 30.69 & 47.19 & 33.57 & 45.35 \\ \midrule
PGDAT & 45.86 & 15.60 & 18.78 & 12.64 & 12.55 & 44.08 & - & 18.24 & 28.06 \\
TRADES ($1/\lambda=1$) & 47.60 & 14.93 & 18.36 & 11.95 & 11.63 & 41.39 & 32.47 & 18.74 & 30.02 \\
TRADES ($1/\lambda=6$) & 43.08 & 19.65 & 23.14 & 17.69 & 14.89 & 44.09 & 31.56 & 18.09 & 26.93 \\
IAT & 54.20 & 17.37 & 20.64 & 13.00 & 10.99 & 41.40 & 38.05 & 24.39 & 34.67 \\ \midrule
AugRmixAT-1-1 & \bf{69.64} & \bf{42.79} & 14.42 & 7.13 & 3.48 & 44.91 & \bf{54.14} & \bf{35.93} & \bf{62.66} \\
AugRmixAT-1-32 & 52.90 & 26.27 &\bf{28.70} &  \bf{22.20} & \bf{16.96} & \bf{50.24} & 39.22 & 24.56 & 42.91 \\

\bottomrule
\end{tabular}
}
\end{table}

\begin{table}[th]
\centering
\caption{Sensitivity of hyperparameters $\lambda_1$ and $\lambda_2$}
\label{tab:lambda}
\scalebox{0.7}{
\begin{tabular}{@{}cc|c|cccc|cc@{}}
\toprule
 &  &  & \multicolumn{4}{c|}{White-box attacks}  &  &  \\
$\lambda_1$ & $\lambda_2$  & Clean & FGSM & PGD10 & PGD20 & CW20 & Corr & Occ \\ \midrule
1 & 1 & 96.91 & 79.81 & 62.43 & 49.23 & 47.12 & 89.28 & 90.16    \\
2 & 1 & 96.89 & 80.66 & 61.20 & 46.95 & 44.41 & 90.34 & 89.01    \\
3 & 1 & 96.95 & 80.42 & 58.84 & 40.69 & 36.88 & 90.27 & 89.85    \\
4 & 1 & 96.94 & 80.63 & 57.99 & 40.33 & 36.74 & 90.64 & 88.65    \\
5 & 1 & 96.82 & 79.88 & 57.43 & 39.00 & 34.69 & 91.05 & 89.33    \\
1 & 2 & 96.34 & 79.68 & 65.20 & 52.78 & 50.78 & 89.46 & 88.98     \\
1 & 4 & 94.33 & 77.26 & 65.12 & 52.59 & 50.21 & 87.55 & 85.87    \\
1 & 8 & 89.94 & 59.73 & 63.16 & 50.70 & 48.19 & 82.50 & 78.10   \\
1 & 16 & 88.05 & 60.03 & 64.18 & 53.17 & 50.06 & 79.93 & 73.88   \\
1 & 32 & 84.81 & 59.61 & 63.66 & 54.86 & 51.83 & 76.41 & 69.52    \\ \bottomrule
\end{tabular}
}
\end{table}
\subsection{CIFAR-100 and Tiny-ImageNet}
We also verify the effectiveness of our method on CIFAR-100 and Tiny-ImageNet. The results are presented in Table~\ref{tab:cifar100}. In Tables~\ref{tab:cifar100}, ``Corr'' is the common corruptions robustness, evaluated using the mean corruption accuracy (mCA$=$$1-$mCE), ``Occ'' is the partial occlusion robustness, evaluated using the mean of Top1 untargeted occlusion robust accuracy and Top2 targeted occlusion robust accuracy. The other settings are the same as on the CIFAR-10.

\subsection{Sensitivity of hyperparameters $\lambda_1$ and $\lambda_2$}
We apply PreAct-ResNet18~\cite{he2016identity} to implement regularization hyperparameters $\lambda_1$ and $\lambda_2$ sensitivity experiments on CIFAR-10. The other settings are the same as the above experiments. We can observe from Table~\ref{tab:lambda} that as the hyperparameters parameter $\lambda_1$ increases, the common corruptions robust accuracy increases while the adversarial robust accuracy decreases. Moreover, as the hyperparameters parameter $\lambda_2$ increases, the clean accuracy, the common corruptions robust accuracy, and the partial occlusion robust accuracy decrease while the adversarial robust accuracy increases. This also verifies that when improving only one specific robustness, it is often detrimental to the robustness of another one or more. In practical applications, we recommend setting both regularization hyperparameters $\lambda_1$ and $\lambda_2$ to 1, which can effectively improve the generalization performance and multiple robustness of the model.

\section{Conclusion}
\label{sec:conclusion}
We propose AugRmixAT, which is a new image data processing and training method. Unlike previous data augmentation and adversarial training, our method not only improves the generalization performance of neural network models but also improves a variety of robustness including white-box robustness, black-box robustness, common corruption robustness, and partial occlusion robustness. Moreover, AugRmixAT can be easily inserted into existing training pipelines, and we believe it can make neural networks used in real-world applications more reliable and secure.

\vfill\pagebreak
\bibliographystyle{IEEEbib}
\bibliography{ref.bib}

\end{document}